\documentclass[sigplan,10pt, screen]{acmart}
\usepackage{tikz}
\usepackage[normalem]{ulem}
\usepackage{amsmath,mathtools}

\usepackage{amssymb}
\usepackage{multirow,booktabs}
\usepackage[linesnumbered,ruled,vlined]{algorithm2e}
\usepackage{float}
\usepackage{graphicx}
\usepackage{subfig}
\usepackage{mathrsfs}
\usepackage{pifont}
\usepackage{booktabs} 
\usepackage{makecell}
\usepackage{paralist}
\usepackage{comment}

\newcommand{\oursys}{{ByteRobust}}
\newcommand{\ourcompany}{ByteDance}
\newcommand{\eg}{\textit{e.g.}}
\newcommand{\ie}{\textit{i.e.}}
\newcommand{\etc}{\textit{etc.}}
\newcommand{\aka}{\textit{a.k.a.}}

\newcommand{\eurus}{StormStance}
\newcommand{\megavision}{EarthStance}
\newcommand{\karl}{FireStance}
\newcommand{\ftruntime}{automatic fault-tolerance runtime}
\newcommand{\analyzer}{aggregation analyzer}
\newcommand{\classifier}{machine status classifier}

\newcommand{\sosp}[1] {\textcolor{black}{#1}}
\newcommand{\hanli}[1] {\textcolor{black}{#1}}
\newcommand{\camera}[1] {\textcolor{black}{#1}}

\AtBeginDocument{%
  }

\begin{document}
\title{Robust LLM Training Infrastructure at \ourcompany{}}

\author{
{Borui Wan$^{1*}$, Gaohong Liu$^{*}$, Zuquan Song$^{*}$, Jun Wang$^{*}$, Yun Zhang$^{*}$, Guangming Sheng$^{1*}$, Shuguang Wang, Houmin Wei, Chenyuan Wang, Weiqiang Lou, Xi Yang, Mofan Zhang, Kaihua Jiang, Cheng Ren, Xiaoyun Zhi, Menghan Yu, Zhe Nan, Zhuolin Zheng, Baoquan Zhong, Qinlong Wang, Huan Yu, Jinxin Chi, Wang Zhang, Yuhan Li, Zixian Du, Sida Zhao, Yongqiang Zhang, Jingzhe Tang, Zherui Liu, Chuan Wu$^{1\dagger}$, Yanghua Peng, Haibin Lin, Wencong Xiao$^\dagger$, Xin Liu, Liang Xiang}
}
\affiliation{
\vspace{2mm}
{\textit{\Large{$^1$The University of Hong Kong}}\hspace{4mm} \textit{\Large{ByteDance Seed}}}
\country{}
}

\thanks{$*$\hspace{1mm}Equal contribution.}
\thanks{$\dagger$\hspace{1mm}Corresponding authors are Chuan Wu <cwu@cs.hku.hk> and Wencong Xiao <hanli.hl@bytedance.com>}

\renewcommand{\shortauthors}{Wan et al.}

\begin{abstract}
The training scale of large language models (LLMs) has reached tens of thousands of GPUs and is still continuously expanding, enabling faster learning of larger models.
Accompanying the expansion of the resource scale is the prevalence of failures (CUDA error, NaN values, job hang, \etc), which poses significant challenges to training stability. Any large-scale LLM training infrastructure should strive for minimal training interruption, efficient fault diagnosis, and effective failure tolerance to enable highly efficient continuous training. 
This paper presents \oursys{}, a large-scale GPU infrastructure management system tailored 
for robust and stable training of LLMs. It exploits the uniqueness of LLM training process and gives top priorities to detecting and recovering failures in a routine manner. Leveraging parallelisms 
and characteristics of LLM training, 
\oursys{} enables 
high-capacity fault tolerance, prompt fault demarcation, and localization with an effective data-driven approach, comprehensively ensuring continuous and efficient training of LLM tasks.
\oursys{} is deployed on a production GPU platform 
and advances the state of the art in training robustness by achieving 97\% ETTR for a three-month training job on 9,600 GPUs. 

\end{abstract}

\ccsdesc[500]{Computer systems organization~Reliability}
\ccsdesc[500]{Hardware~Fault tolerance}
\ccsdesc[500]{Computing methodologies~Distributed artificial intelligence}
\ccsdesc[500]{Computing methodologies~Natural language processing}

\keywords{LLM Training, Fault Tolerance, Fault Diagnosis}

\acmYear{2025}\copyrightyear{2025}
\setcopyright{cc}
\setcctype[4.0]{by}
\acmConference[SOSP '25]{ACM SIGOPS 31st Symposium on Operating Systems Principles}{October 13--16, 2025}{Seoul, Republic of Korea}
\acmBooktitle{ACM SIGOPS 31st Symposium on Operating Systems Principles (SOSP '25), October 13--16, 2025, Seoul, Republic of Korea}
\acmDOI{10.1145/3731569.3764838}
\acmISBN{979-8-4007-1870-0/25/10}

\maketitle


\section{Introduction}
\label{sec:intro}

Large Language Models (LLMs)~\cite{gpt4,llama3.1,google-gemini}  have elevated generative artificial intelligence to an unprecedented level.
They are being applied in a wide range of domains such as 
chatbots~\cite{chatgpt, claude} and programming assistants~\cite{copilot, code-llama}, spawn various speculation on future applications~\cite{openai-o1, llm-game}, and are 
exerting profound impacts on people's life and work styles.

Training an LLM demands huge amounts of resources and long training time. The pretraining of LLaMA 3, a 405-billion-parameter model, involves 16,384 NVIDIA H100 GPUs and spans 54 days~\cite{llama3.1}. A 175-billion-parameter model was trained in ByteDance using 12,288 GPUs~\cite{megascale}.
Recently, xAI established a cluster of 100,000 GPUs to further scale training~\cite{xai100k}. Given the massive scale of distributed training over extended durations, failures (\eg, CUDA error, NaN values, job hang, \etc) are nearly inevitable.
Meta reports that hardware failures occur approximately every 2.78 hours during the training of large models on 16,000 GPUs~\cite{llama3.1}. 

\camera{
For LLM training, current failure diagnosis and handling practice typically relies on log analysis and exit code evaluation following the fail-stop events~\cite{megascale, characterization}, or monopolizing the entire cluster to conduct stress testing~\cite{superbench}.
}
Once the root cause is identified, the training job is resumed with rescheduled resources and parallel configuration~\cite{meta-cluster}, 
reloading checkpoints, often consisting of terabytes of data, 
from a remote file system~\cite{checkfreq, check-n-run, bytecheckpoint}. 
This fail-stop, diagnosis and reassuming procedure incurs non-negligible overhead, ranging from several hours to even days~\cite{unicron},
which constrains the effective training time ratio (ETTR, calculated as the ratio between the productive training time and the wall-clock time of a job) in the face of high fault frequencies that increase with the expansion of model/resource scales.


\hanli{There are many factors contributing to the instability of training large-scale models. Firstly, we have observed that many errors do not manifest as \textit{explicit failures}, \ie{}, failures where the fault source can be accurately located through error messages and faulty machines. Instead, they include numerous \textit{implicit failures}. Such implicit failures encompass scenarios where the fault itself lacks clear signals, such as jobs hanging without making progress, training trajectories deviating unexpectedly, performance jittering and degrading, as well as hard-to-locate root-cause failures like silent data corruption~\cite{sdcgoogle, sdcscale, llama3.1, sdcllm} (SDC), \eg, NaN loss values.}
Current approaches often rely on timeouts and termination of training processes to identify faulty machines~\cite{characterization, unicron}, which may incur significant waste of GPU cycles. MegaScale~\cite{megascale} uses plummeting RDMA traffic as an indicator of implicit errors, and still requires manual investigation to fully identify 
the issue and pinpoint the root causes.

\hanli{Secondly, the ultra-large training scale also poses challenges to sustained, stable training.
When scaling to tens of thousands of GPUs, even if a current failure is identified as caused by some machines, there are not enough spare machines to replace all training resources for recovery, making the localization and isolation of faulty machines a critical path in the entire stable training process.
Therefore, compared to stable fault-tolerant training in small-scale task scenarios, the expansion of scale not only increases the frequency of failure interruptions but also introduces more difficult phases, making the system even more challenging.}


\hanli{Lastly, compared to small-scale and short-duration model training, the months-long LLM pretrain presents a different paradigm shift, namely \textit{the continuous evolution of user code}, with ongoing integration of performance optimizations or algorithmic adjustment updates.}
It arises primarily from the unprecedented scale, complex parallelization strategies, and the pursuit of near-optimal resource utilization (\eg, GPU memory) to save LLM training costs.
Optimization techniques and parallelization strategies derived from small-scale testing~\cite{pipedream, megatron2021, ringattention, ulysses, zerobubble, gshard, flux2024} are often sub-optimal at large scales; continuous optimization based on profiling during training job execution becomes essential. Additionally, the extended duration of model training, often spanning months, frequently necessitates the integration of new versions of code and modules. All these require frequent job interruptions and restarts, introducing additional complexity to maintaining stable distributed training. Furthermore, the evolving code itself can be a source of errors, particularly those undetectable during small-scale testing that fail to simulate real production environments.

\camera{Driven by in-production observations and experience (details in Sec.~\ref{sec:motivation}), we build \oursys{}, a robust LLM training infrastructure that achieves our key goal: \textit{highly efficient incident diagnosis and handling with minimal unproductive time.}}
\oursys{} is carefully designed to monitor and manage the entire lifecycle of LLM training~\cite{llama3.1, wei2021sft, wang2022sft, hybridflow} to handle training incidents automatically and efficiently at scale.
Unlike conventional GPU management and fault tolerance systems~\cite{gandiva2018, megascale,characterization,aws-gemini} which typically operate at Kubernetes pod levels, \oursys{} extends LLM training job manifests to include fine-grained process management, capable of leveraging runtime information for failure detection and achieving fast recovery.
\camera{
\oursys{} achieves this goal with a comprehensive set of techniques, based on our novel system design philosophy as summarized below.
}

\noindent \textbf{\camera{Prioritize rapid isolation, not precise localization.}}
\sosp{
\oursys{} favors rapid fault isolation over exhaustive localization.
In large‐scale LLM training (often spanning thousands of GPUs), precise failure pinpointing can leave vast GPUs idle~\cite{megascale, llama3.1}.
To maximize ETTR, we combine lightweight real‐time detection with hierarchical stop‐time diagnostics, quickly singling out faulty machines with minimal overhead.
When these approaches fall short, \oursys{} applies data‐driven clustering of runtime stack-traces to isolate suspect machines at defined fault domains, i.e., parallel groups, over-evicting them rather than chasing exact root causes.
}

\noindent \textbf{\camera{Incorporate human error in design.}}
\sosp{
Unlike standard DL training jobs~\cite{optimus, gandiva2018}, multi‑month LLM training entails continuous updates to data, algorithms, and engineering, which compounds system vulnerability.
Recognizing the human error as an inevitable failure source, \oursys{}'s automated fault tolerance framework combines machine fault detection and diagnostics with code rollback 
for rapid verification and recovery.
\hanli{Further, user code changes are merged with deterministic failures through a lazy update approach, utilizing the inevitability and high frequency of failures.}
}

\noindent \textbf{\camera{Control variability during swift recovery.}}
\sosp{
Failures stem from both hardware faults and software bugs, and machines can degrade over long‐running jobs.
Ensuring stability during code upgrades and recovery is therefore essential.
For changes that don’t alter machine allocation, we use an in-place hot‑update mechanism to preserve the runtime environment and simplify diagnostics.
\camera{
To ensure controlled and rapid recovery, \oursys{} leverages pre‑provisioned warm standbys that execute self-checks before delivery to avoid full job rescheduling.
}
Finally, our checkpointing module dovetails with the failure domains by distributing backups across parallel groups, outside any single failure domain, eliminating remote‐fetch dependencies to achieve rapid restarts.
}

\camera{
\oursys{} has been deployed for over one year to facilitate our in-house LLM training in high-performance production GPU clusters. 
}
\oursys{} identifies 38,236 explicit failures, 5,948 implicit failures in the three-month period through the automated fault tolerance training framework.
Our micro-benchmark experiments on 16,384 GPUs show that the warm standby and hot update mechanisms achieve up to $10.87\times$ and $11.04\times$ respectively in recovery.
The efficient checkpoint mechanism in \oursys{} achieves every-step checkpointing with less than 0.9\% overhead to speed up failover.
Deployment experiments show \oursys{} achieves up to 97\% ETTR for a three-month 9,600 GPUs training job.

\vspace{-2mm}
\section{Background and Motivation}
\label{sec:motivation}

\subsection{Characteristics of LLM Training}
\label{sec:llm_training}

\begin{figure}[!t]
  \centering
  \includegraphics[width=\linewidth]{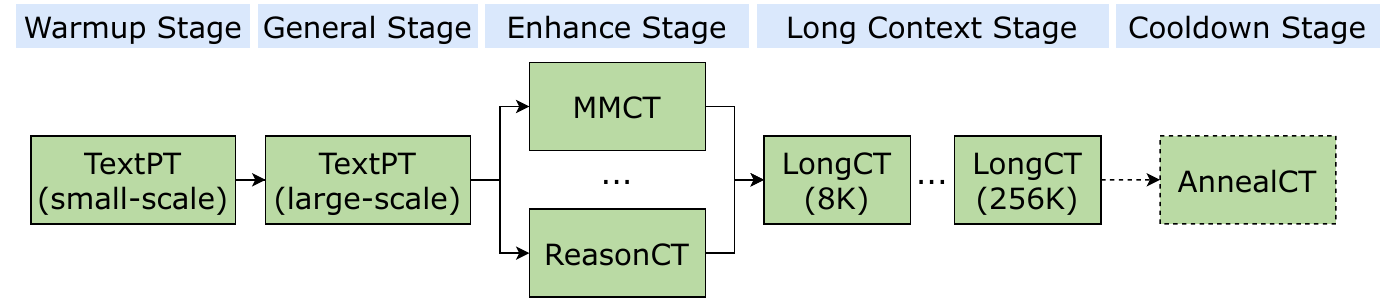}
  \caption{
  \camera{
  Recipe of LLM pretraining.
  TextPT: Text Pretraining; MMCT: Multimodal Mixed Continual Training; ReasonCT: Reasoning Continual Training; LongCT: Long Context Continual Training; AnnealCT: Annealing Continual Training.
  Different LLMs may reorder stages~\cite{kimi-k2, llama3.1}.
  }
  }
  \label{fig:multi-stage}
\end{figure}

\begin{figure}[!t]
  \centering
  \includegraphics[width=\linewidth]{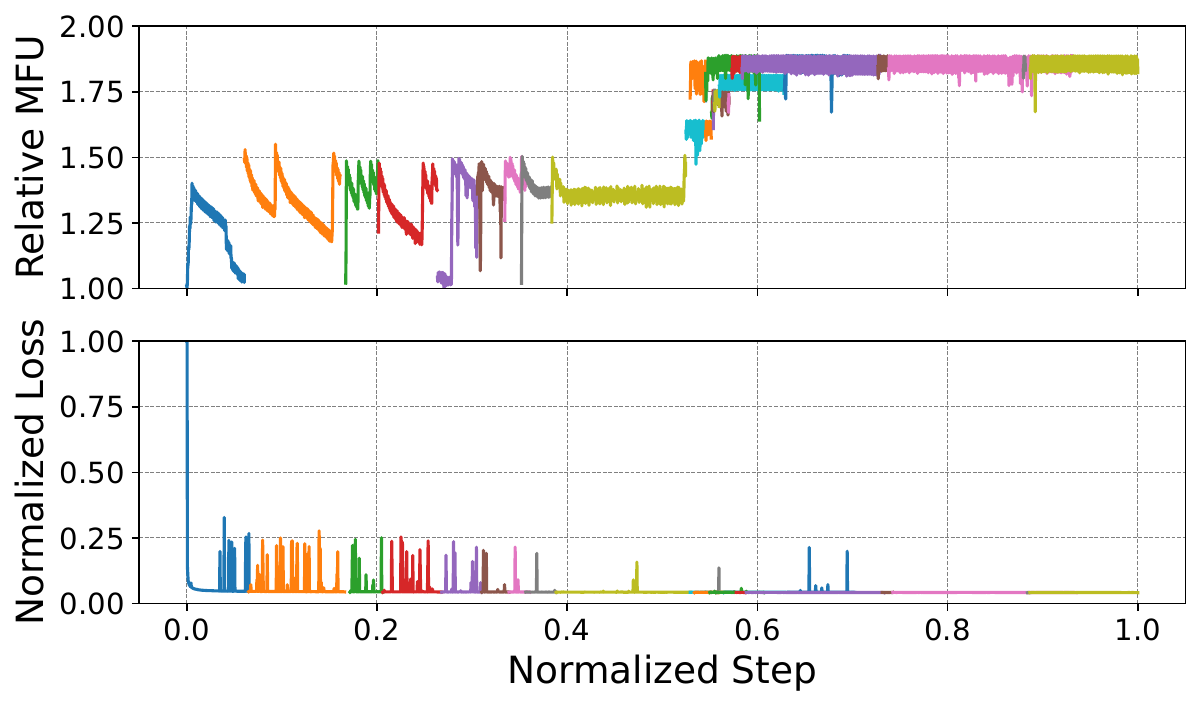}
  \caption{
  Normalized loss and relative MFU (ratio to the minimum MFU value) curves of an LLM training job running on 1000 GPUs in a production environment.
  Each color indicates one continuous, uninterrupted training period. 
  }
  \label{fig:freq_restart}
\end{figure}

\noindent \textbf{Complex parallelism strategies.}
A variety of parallelism strategies have been exploited in distributed LLM training, including data parallelism (DP)~\cite{pytorch-distributed, fsdp}, tensor parallelism (TP)~\cite{megatron2019}, pipeline parallelism (PP)~\cite{gpipe, pipedream, megatron2021, zerobubble}, and sequence parallelism (SP)~\cite{ulysses, ringattention}. Gradient checkpointing~\cite{grad-checkpointing} and CPU offloading~\cite{zero-offload} are often integrated 
for GPU memory optimization.
Adam optimizers~\cite{adam} consume $6\times$ GPU memory compared to model weights~\cite{characterization}. Zero Redundancy Optimizer (ZeRO)~\cite{zero} is adopted to reduce the memory footprint by sharding the optimizer states (ZeRO-1), gradients (ZeRO-2), and model parameters (ZeRO-3) across GPUs.

\noindent \textbf{\camera{Multi-stage configuration adjustments.}}
\camera{
Whereas traditional DL jobs often run with a single, fixed configuration and unchanging user code~\cite{gandiva2018, optimus}, LLM pretraining unfolds over multiple stages, each demanding shifts in both algorithmic paradigm and system optimization.
Consequently, user code must evolve continuously to meet differing optimization goals.
Fig.~\ref{fig:multi-stage} illustrates a typical five-stage LLM pretraining pipeline~\cite{llama3.1, deepseek-v3, kimi-k2, qwen3, seed-1.6}:
(i) \textit{Warmup Stage}.
A small-scale pure text pretraining runs with a reduced DP size to validate algorithmic changes.
Frequent code tweaks here ensure stability and early performance gains~\cite{bytecheckpoint}.
(ii) \textit{General Stage}.
Full-scale pretraining on a broad text corpus to absorb knowledge.
Since training scales up, engineering codes are iteratively refined for optimal throughput and memory efficiency~\cite{megascale}.
(iii) \textit{Enhance Stage}.
Data mixtures are re-weighted to bolster specific capabilities, \eg, high-quality STEM, coding, and math datasets for improved reasoning~\cite{qwen3} or multimodal corpora for cross-modal understanding~\cite{kimi-k2, seed-1.6}.
Apart from data adjustments, user code may also be extended to incorporate additional adaptors~\cite{dosovitskiy2021image} or modality-specific optimization techniques~\cite{seed1.5-vl, overlord}.
(iv) \textit{Long Context Stage}, the context windows and the allocated machines are progressively expanded (\eg, from 8K to 256K).
Scenario-tailored engineering codes such as Hybrid Data Parallelism (HDP)~\cite{bytescale} are integrated to sustain efficiency at extreme sequence lengths.
(v) optional \textit{Anneal Stage}, certain domain-specific or synthetic datasets are carefully unsampled~\cite{llama3.1, kimi-k2} to tune and stabilize the final performance.
}

\begin{table}[!t]
\caption{Statistics of training incidents collected over a three-month span, encompassing 778,135 LLM training jobs.
}
\label{tab:incident_stat}
\resizebox{\linewidth}{!}{
\begin{tabular}{c|c|c|c}
\toprule
\textbf{Category}                   & \textbf{Incident Symptom}                 & \textbf{Count} & \textbf{Percentage} \\
\midrule
\multirow{12}{*}{Explicit} & CUDA Error                        & 19968 & 36.1\%     \\
& CPU Overload                      & 6095  & 11.0\%     \\
& CPU OOM                           & 5567  & 10.1\%     \\
& Insufficient Disk Space           & 2755  & 5.0\%      \\
& Infiniband Error           & 1599  & 2.9\%      \\
& Filesystem Mount                 & 1176  & 2.1\%      \\
& HDFS~\cite{hdfs} Error                        & 1104  & 2.0\%      \\
& Container Error                   & 781   & 1.4\%      \\
& OS Kernel Panic                   & 203   & 0.4\%      \\
& GPU Memory Error                  & 188   & 0.3\%      \\
& External Service Error       & 128   & 0.2\%      \\
& GPU Unavailable                   & 76    & 0.1\%      \\
& Disk Fault                           & 47    & 0.1\%      \\
\midrule
\multirow{2}{*}{Implicit}  & Job Hang                         & 5506  & 9.9\%     \\
& MFU Decline                       & 442   & 0.8\%      \\
& NaN value                          & 148   & 0.3\%      \\
\midrule
Manual Restart                & Code/Data Adjustment             & 9582  & 17.3\%   \\ 
\bottomrule
\end{tabular}
}
\end{table}

\begin{figure*}[!t]
  \centering
  \includegraphics[width=0.95\linewidth]{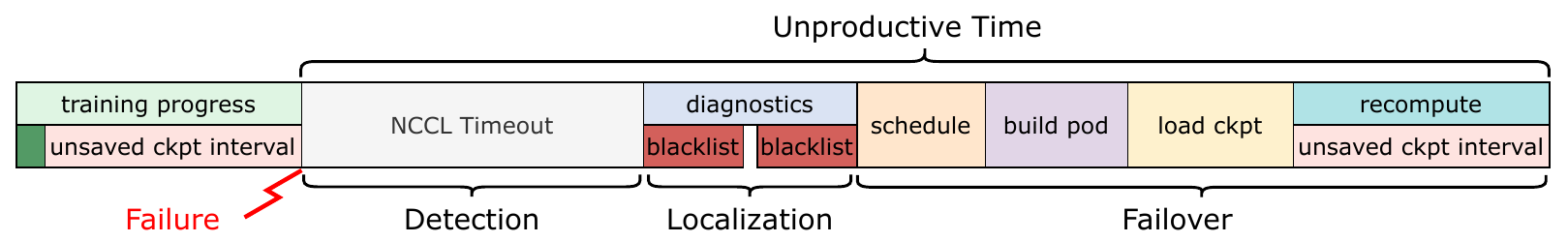}
  \caption{
  \camera{
  Unproductive time breakdown upon failures.
  Implicit failures, such as job hangs, are selected as examples since they typically result in prolonged unproductive times.
  }
  }
  \label{fig:unproductive_time}
  \vspace{-3mm}
\end{figure*}

\noindent \textbf{Frequent failures and restarts.}
Large-scale LLM training is often plagued by frequent failures and restarts.
Fig.~\ref{fig:freq_restart} shows the training loss and Model FLOPs Utilization (MFU) when training an LLM in a 1000-GPU cluster over a 10-day training span, during which a total of 28 runs were conducted (each run corresponds to a restart of the model training). The interruptions were caused by infrastructure 
issues, as well as manual adjustments to the training algorithm and strategies. 
As training progresses, the loss gradually decreases while the MFU increases, reflecting the impact of engineering efforts such as tuning parallel strategies, integrating fused computational kernels, \etc,
exerted for improving performance upon each restart.
Notably, when a manual restart occurs, the training progress might be intentionally rolled back by several steps to verify the correctness of the engineering improvements and to ensure that the loss curves remain bitwise consistently aligned, which corresponds to the curve overlap observed in some of the runs.

\subsection{Observations on Training Incidents}
\label{sec:observe}

\noindent \textbf{Incident distribution.}
Table~\ref{tab:incident_stat} summarizes training incidents in all LLM training jobs captured on our production platform over the last three months.
We classify these incidents into three categories, based on their characteristics: 
(i) \texttt{Explicit failures}, which are characterized by clear diagnostic indicators, such as error messages in \textit{stdout} or \textit{stderr} logs or specific exit codes; (ii) \texttt{Implicit failures}, which often manifest as job hang-ups, performance degradation, or anomalous training trajectories,
whose root causes are often elusive; 
and (iii) \texttt{Manual restart}, which refers to proactive interruptions of training for the purposes of algorithm and engineering improvement. For instance, optimization techniques such as kernel fusing~\cite{flux2024}, computation-communication overlapping~\cite{centauri}, \etc, are continuously integrated; updates to software versions and adjustments in parallelism configurations are also made to enhance training efficiency. 
To mitigate undesirable trajectories like loss spikes, changes in algorithmic code are also necessary, \eg, skip problematic mini-batches,  tune hyper-parameters~\cite{gpt3, palm, glm}.

\noindent \textbf{\camera{Different unproductive times.}}
\camera{
The breakdown of unproductive time caused by training incidents is shown in Fig.~\ref{fig:unproductive_time}, which includes detection, localization, and failover times.
}
\sosp{
As the majority of incidents, explicit failures enable immediate faulty machine identification or application of targeted diagnostics~\cite{superbench, characterization, megascale, aegis}.
These failures typically exhibit short detection times (around 60s, \camera{via error messages or other log-based indicators}) and localization times (ranging from 2 minutes to 15 minutes).
In addition, more than 10\% of incidents manifest as implicit failures and are difficult to detect or pinpoint their sources.
For instance, we encountered a communication hang issue caused by CUDA errors (see this example in Sec.~\ref{sec:agg_analysis}), which took us more than one and a half hours for manual diagnosis.
}
\camera{
Besides, SDC~\cite{sdcgoogle, sdcscale, llama3.1, sdcllm} on GPU hardware has emerged as a critical challenge in LLM training.
It appears as stochastic faults, such as abrupt loss divergence or NaN values, and often makes stop-time diagnostics unable to reproduce within a short period of time.  
We uncovered a data-type-dependent computation error rooted in GPU SDC, characterized by non-deterministic behavior that hindered reliable reproduction.
Through more than 8 hours of offline stress testing, we successfully identified the faulty GPU machines.
}
Finally, as LLM training is akin to a scientific experiment, $9582$ manual interruptions occurred, primarily due to modifying the execution code to optimize training performance or to test new configurations.
\camera{
The unproductive time for them includes only failover time.
}

\subsection{\camera{Challenges to Achieve High ETTR}}
\label{sec:challenges}

\begin{table}[!t]
\caption{
Root cause of incidents
}
\label{tab:cross_domain_distribution}
\begin{small}
\resizebox{\linewidth}{!}{
\begin{tabular}{cccc}
    \toprule
    \textbf{Symptom} & \textbf{\#Infrastructure} & \textbf{\#User Code} & \textbf{\#Total} \\
    \midrule
    Job Hang & 21 & 5 & 26 \\
    Illegal memory access & 21 & 41 & 62 \\
    NaN value & 3 & 1 & 4 \\
    \bottomrule
\end{tabular}}
\vspace{-3mm}
\end{small}
\end{table}

Effectively mitigating diagnosis and recovery overhead of various failures and restarts is crucial to ensure training efficiency of large-scale LLMs, but is complicated as follows.

\noindent \textbf{Complex root causes of failures.}
Under the same symptoms, the root cause of failures can be 
tangled among different aspects. 
In Table~\ref{tab:cross_domain_distribution}, we summarize incidents in the large-scale training jobs (on >2000 GPUs) \sosp{over the last month},
exhibiting three typical symptoms and categorize their root causes into two types:  
infrastructure and user code. 
Failure categories as \textit{infrastructure} are raised from issues within the underlying hardware or software,  such as GPUs or remote storage~\cite{characterization}.
\textit{User code} failures typically stem from programming or configuration errors in model development or training frameworks by engineers.
Job Hang
can be triggered by NVIDIA IB Switch Unified Fabric Manager (UFM) faults (infrastructure issue) or mis-configurations of checkpoint resharding (user code issue). 
GPU memory errors like \texttt{illegal memory access} can be caused by the broken High Bandwidth Memory (HBM) or incorrect implementation of computation kernels in handling variable sequence lengths, i.e., both infrastructure error and user code error are possible.
\sosp{
Furthermore, NaN values (\eg, loss NaN) can also stem from multiple sources, including problematic data, code bugs, or hardware-induced SDCs.
This multi-factorial origin makes it hard to identify the fault sources, particularly in distributed training~\cite{nan2023}.
}
Log analysis based on specific rules can hardly diagnose failure from different aspects automatically.
User code evolution further exacerbates the difficulty of fault localization. 
When a new code version is applied, the running job is stopped, and the new job is launched. 
If a failure occurs at this moment, it is difficult to distinguish whether the root cause is due to user code or the infrastructure.

\noindent \textbf{Implicit failures are hard to detect and locate.} According to Table~\ref{tab:incident_stat}, 
a typical indicator of implicit failures is \texttt{Hang}, covering over 10\% of all incidents. 
Existing LLM robust training systems rely on log analysis to monitor the training health status~\cite{characterization, unicron}; 
when hang happens, the training job will not produce any logs until timeout (\ie, 30 or 60 minutes for NCCL). 
Utilizing RDMA traffic monitoring,
MegaScale~\cite{megascale} can identify anomalous behavior earlier, but locating the root-cause anomalous machines or GPUs is still challenging. 
When MFU declines or 
fluctuates, all measurements regarding 
IO, computation, and communication decrease or fluctuate simultaneously, challenging root cause diagnosis.
\camera{
Apart from job hang and performance degradation, SDC is also hard to handle.
This issue is further compounded by the collective communication paradigm intrinsic to distributed LLM training, which exacerbates SDC propagation: a single corrupted gradient from one node can contaminate the global parameter update across all participating workers.
This amplifying effect is particularly insidious because it obfuscates the original fault locus within the aggregated signals.
}

\noindent \textbf{\camera{Uncertainty and high overhead of failover.}}
\camera{
Upon fault isolation or manual adjustments, failover is triggered for training recovery.
As depicted in Fig.~\ref{fig:unproductive_time}, failover operations include: scheduling new machines~\cite{meta-cluster} associated with the terminated job, reconstructing the pod environment, reloading the latest checkpoint from the remote storage~\cite{bytecheckpoint}, and recomputing the lost training progress.
During the machine rescheduling process, degraded machines may be assigned, introducing potential new faults after job restarts.
For manual restarts that include code upgrades, this process can create ambiguity in determining whether new faults stem from the recently integrated code or the underlying infrastructure.
In addition to its uncertainty, failover is inherently costly, leading to significant unproductive time.
For instance, retrieving checkpoints from remote storage over low-bandwidth frontend networks can be notably time-consuming.
Furthermore, relying on remote storage leads to substantial recomputation overheads, as relatively large checkpointing intervals (e.g., 30 minutes~\cite{check-n-run} or every 100 training steps~\cite{bloom, bytecheckpoint}) are required to reduce checkpoint stalls.
We observed that the time cost of failover operations is often more than 10 minutes for the large model training at the scale of 10,000 GPUs.
As the scale of LLM training increases, the failure frequency increases~\cite{llama3.1, glm, opt}, amplifying the overhead of failover operations in the entire training span.
}

\vspace{-1mm}
\section{\oursys{} Overview}
\label{sec:design}

\begin{figure}[!t]
  \centering
  \includegraphics[width=\linewidth]{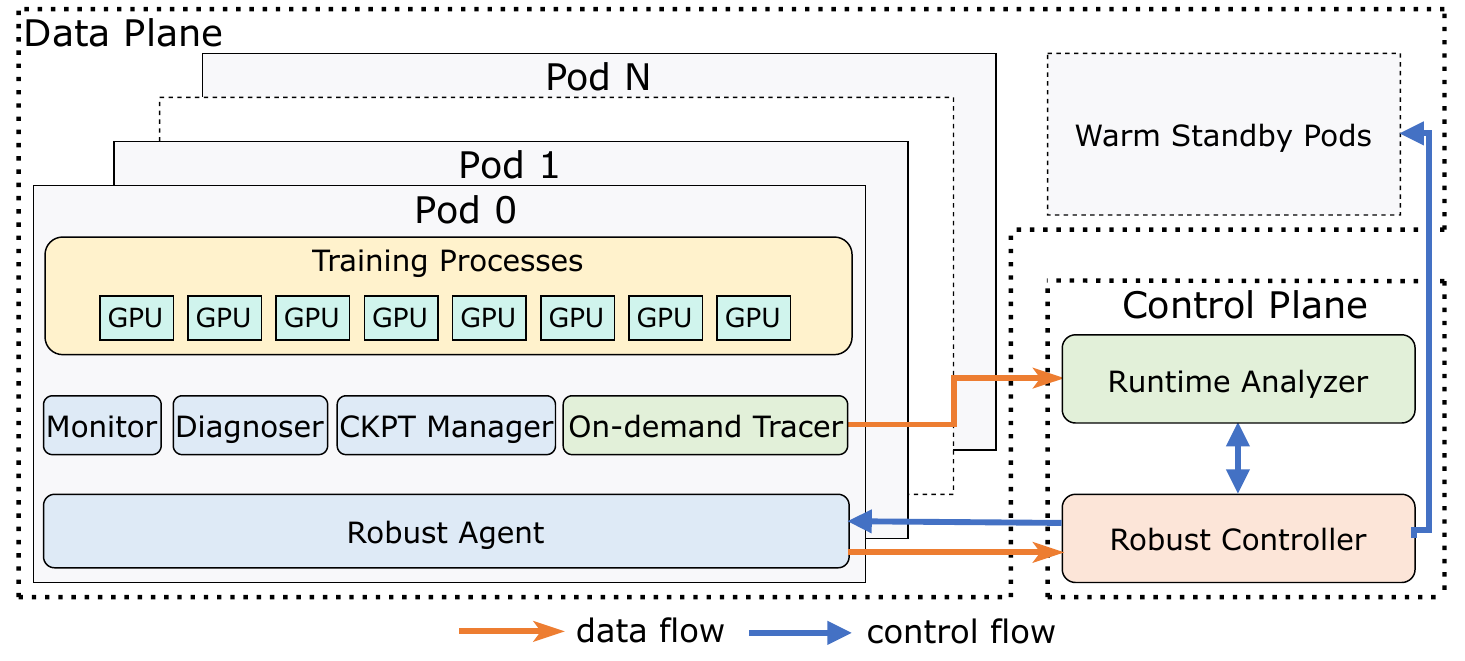} 
  \caption{
  Architecture of \oursys{}.
  }
  \label{fig:arch}
  \vspace{-6mm}
\end{figure}

\begin{figure*}[!t]
  \centering
  \includegraphics[width=\linewidth]{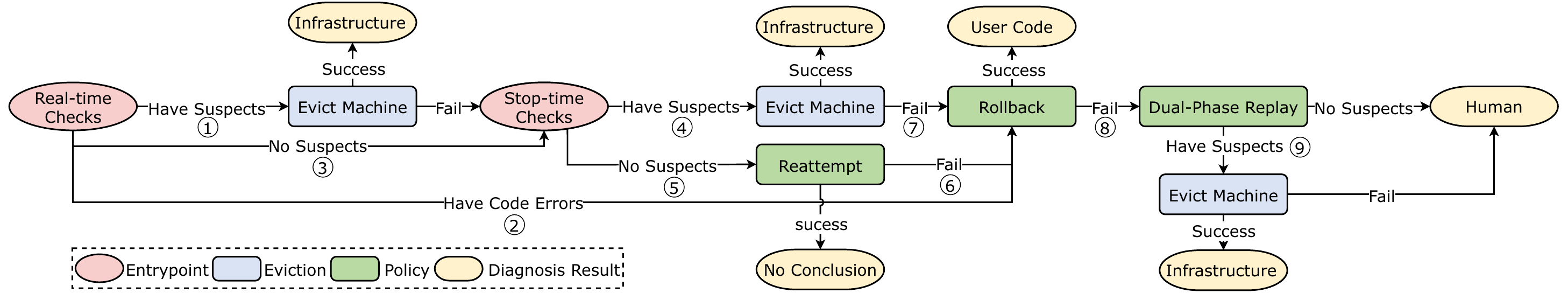}
  \caption{
  The automated fault tolerance mechanism of \oursys{}. 
  }
  \label{fig:recovery_workflow}
\end{figure*}


\camera{
\oursys{} is designed to address the above challenges and fulfill our primary goal of achieving high ETTR in large-scale LLM training:
automatically diagnosing and addressing various training incidents while minimizing unproductive time.
As illustrated in Fig.~\ref{fig:arch}, \oursys{} consists of two core components, \textit{control plane} and \textit{data plane}.
The control plane operates external to the training job, orchestrating the robust incident-handling strategy that detects anomalies, localizes faults, and triggers appropriate recovery actions.
The data plane resides within each training pod and integrates modules for monitoring, diagnosing, checkpoint management, and stack-trace capture, furnishing real-time observability, immediate diagnostics upon interruption, rapid checkpoint rollbacks, and on-demand aggregation analysis.
}

\noindent \textbf{Control plane.}
\camera{
The control plane comprises two modules that enable robust failure detection, localization, and recovery in LLM training.
\textit{Robust Controller} orchestrates an automated failure mitigation framework (Sec.~\ref{sec:auto_ft}), leveraging real-time monitoring and stop-time diagnostics to handle most incidents.
For controlled and swift recovery, it uses an in-place hot-update mechanism to restart training when no machine is evicted (Sec.~\ref{sec:hot_update}).
When certain machines are decided to be evicted, it requests warm standby machines, pre-validated via self-checks, to resume the job (Sec.~\ref{sec:warm_standby}).
\textit{Runtime Analyzer} addresses job hangs and performance degradations by aggregating stack traces from training pods to isolate and (over-)evict suspected machines (Sec.~\ref{sec:analyzer}).
}



\noindent \textbf{Data plane.}
\camera{
The \textit{Robust Agent} daemon runs in each training pod, processing control signals from the Robust Controller and managing four sub-modules below:
\textit{Monitor} collects multifaceted data to detect outliers, supporting real-time checks (Sec.~\ref{sec:proactive_checks}) and triggering aggregation analysis upon anomalies.
\textit{Diagnoser} runs domain-specific benchmarks and test suites~\cite{superbench, testyuan} after job suspension, enabling in-depth diagnosis of complex failures (Sec.~\ref{sec:hier_checks}).
\textit{On-Demand Tracer} captures stack-traces from training processes (when aggregation analysis is invoked) and uploads them to the Runtime Analyzer.
\textit{CKPT manager} performs asynchronous checkpointing with cross-parallel-group backups to CPU memory and local disk, minimizing resuming costs (Sec.~\ref{sec:checkpoint}).
}

\section{Automated Fault Tolerance}
\label{sec:auto_ft}

\camera{
Automated fault tolerance is essential for scaling LLM training.
By detecting, localizing, and resolving incidents with minimal human intervention, it dramatically reduces unproductive time.
Besides, since GPU cycles are the most expensive resources in a training cluster, rapid, coarse-grained fault isolation often yields a better trade-off than expensive, fine-grained root-cause pinpointing between diagnostic coverage and efficiency.
To meet these requirements, we propose an automated fault tolerance framework (Fig.~\ref{fig:recovery_workflow}) that combines real-time checks for immediate detection of common errors, stop-time diagnostics for in-depth analysis of complex failures, in-place reattempts to recover from transient faults, code rollbacks to revert from defective user codes, and replay tests to address corner cases such as SDCs.
}

\subsection{Proactive Real-time Checks}
\label{sec:proactive_checks}




\noindent \textbf{System inspection.}
The monitor employs inspection threads to carry out a series of \sosp{lightweight} system health status queries 
at predefined second-level intervals.
These inspections introduce no workload to GPUs and are transparent to the ongoing training job.
The inspections mainly cover:
(i) \textit{Network-side items}, such as NIC down or jitter, 
packet loss rate, 
switches down. 
(ii) \textit{GPU-side items}, including the status of DCGM service~\cite{DCGM}, 
PCIe bandwidth, memory row remapping~\cite{remap}, 
and GPU temperature, \etc
(iii) \textit{Host-side items}, such as OS kernel event (\eg, Xid~\cite{xid} in \texttt{dmesg}).
We set different inspection intervals \sosp{and triggering thresholds for these items to tolerate automatic recovery.
}

Once any anomaly is detected (step $\textcircled{1}$), the monitor reports the warning event to the robust agent, which then notifies the robust controller.
For highly confident events that point to specific machines, such as GPU Unavailable, Disk Fault, the controller stops all processes immediately and evicts problematic machines, skipping stop-time diagnostics (Sec.~\ref{sec:hier_checks}).
For \sosp{network issues},
the controller tolerates several alerts (\eg, twice within 5 minutes empirically) before evicting problematic machines, since some of them (\eg, NIC and Network Switch flipping) can automatically recover themselves~\cite{c4}.
If restarted training fails again after machine eviction, \oursys{} enter the stop-time check procedure.


\noindent \textbf{Metrics collection.} 
The monitor also gathers various metrics 
based on three categories of data: 
(i) \textit{workload-specific training metrics}, including loss, gradient norm, MFU, \etc\
\camera{
We leverage wandb~\cite{wandb} to collect these continuously observable metrics, considering significant changes in them as the faulty signals, \eg, 5$\times$ increase in loss/gradient norms, NaN values.
}
(ii) \textit{stdout/stderr logs and process exit codes}, which serve as hints for diagnostics.
(iii) \textit{events}, including CUDA, RDMA, host, and storage events.
These events are crucial for deriving system performance metrics such as RDMA traffic and TensorCore utilization.
Given the periodic nature of LLM training, significant declines in these metrics serve as a signal for potential 
\sosp{job hangs and MFU declines.}

\sosp{
During runtime, the controller analyzes collected metrics.
If it detects user‑space errors, \eg, TypeError, IndexError, traceable to specific code modules from logs and exit codes, it triggers a code rollback (step $\textcircled{2}$).
If \camera{training crashes} or abnormal metrics, \eg, NaN losses, arise without a clear culprit, it suspends training and runs stop‑time checks (step $\textcircled{3}$).
On spotting performance anomalies, \eg, zero RDMA traffic \camera{within 10 minutes} or low TensorCore utilization, the aggregation analysis is triggered for machine isolation (Sec.~\ref{sec:analyzer}).
}



\begin{algorithm}[!t]
\caption{
\sosp{Dual-Phase Replay} 
}
\label{alg:dual-group}
\SetKwInOut{Input}{Input}
\SetKwInOut{Output}{Output}
\SetKwProg{Fn}{Function}{}{}

\Input{
    $z$: Total number of machines \\
    $m$: Group size (recommended as PP size multiple) \\
    $n \gets  z/m $: Number of groups
}
\Output{Suspect set $S \subseteq \{0,1,...,z-1\}$}


\BlankLine

\Fn{LocateFaultyMachines($z$, $m$, $n$)}{
    The machine $i$ is assigned the ID $x_i$

    Phase 1: Horizontal Grouping \\
    \Indp
    Partition machines into $ n $ groups by $x_i/m$ \\
    Identify faulty group $a$ via replaying \\
    \Indm
    
    Phase 2: Vertical Grouping \\
    \Indp
    Re-partition into $n$ groups by $x_i \bmod n$ \\
    Identify faulty group $b$ via replaying\\
    \Indm
    
    Solve the constrained system:
    \begin{align*}
        \left\lfloor \frac{x_i}{m} \right\rfloor &= a \\
        x_i \bmod n &= b
    \end{align*}
    
    Determine solution cardinality:
    \[
    |S| = \begin{cases}
        1 & \text{if } m \leq n \\
        \lceil m/n \rceil & \text{otherwise}
    \end{cases}
    \]
    
    \Return $S$;
}

\end{algorithm}

\vspace{-5mm}
\subsection{Hierarchical Stop-time Checks}
\label{sec:hier_checks}

\camera{
Although proactive real-time checks can utilize inspections to connect most of the explicit failures to faulty machines, there still exist some errors that are hard to resolve with only real-time collected information.
\oursys{} takes potential human errors into consideration and conducts hierarchical stop-time checks to handle these cases.
}

\noindent \textbf{Diagnose.}
The diagnoser analyzes the logs and exit code for failure diagnosis, running corresponding tests to locate the root causes.
For instance, upon \texttt{NCCL Internal Error},  
NCCL tests are conducted: It first runs NVIDIA Extended Utility Diagnostics (EUD)~\cite{eud} to confirm whether there are obvious errors in GPUs. If no, an intra-machine \textit{all-to-all} test is run to verify if inter-GPU connection bandwidth meets expectations.
If the intra-machine test is passed, the inter-machine machine communication test is conducted.
Each machine runs an \textit{all-gather} test with neighboring machines to verify the connectivity and integrity of data transfer.
The discovered suspected machines are evicted with their IP addresses blocked (step $\textcircled{4}$).
After that, warm standby machines are awakened to restart training (Sec.~\ref{sec:warm_standby}). 

\noindent \textbf{Reattempt.}
If all tests are passed, the diagnoser assumes that the failures are caused by transient faults such as temporary link flapping, switches down, connection reset, \etc\
\camera{The training job is then directly restarted (step $\textcircled{5}$).}

\noindent \textbf{Rollback.}
When restarting training fails to resolve the problem (step $\textcircled{6}$) or training crashes again after machine eviction (step $\textcircled{7}$), the diagnoser assumes that recent updates of user code are highly risky.
It then rollbacks the user code \sosp{with the hot-update mechanism (Sec.~\ref{sec:hot_update})} to remove integrated new features (\eg, newly fused computational kernels) and restarts training.
If training restarts successfully, the user code is deemed as the root causes.
Relevant teams are involved to examine the reliability of their codes while training keeps progressing.

\begin{figure}[!t]
  \centering
  \includegraphics[width=0.95\linewidth]{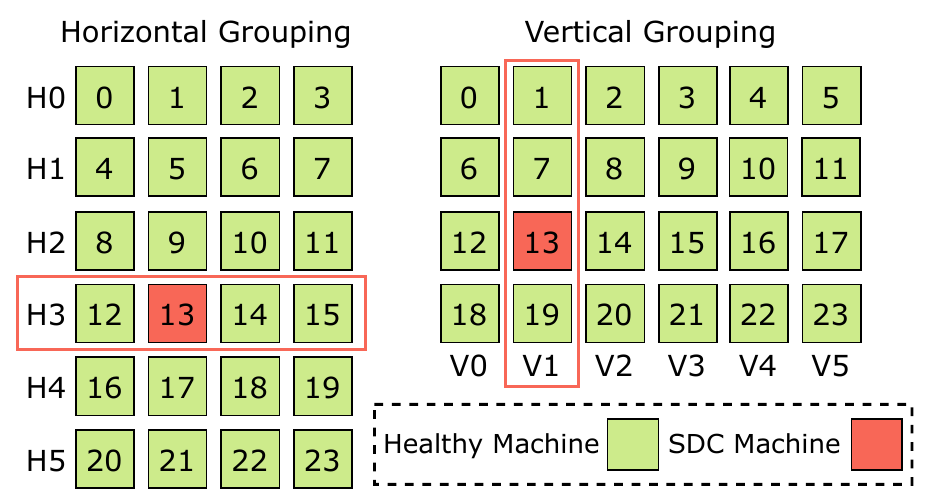}
  \caption{
  \camera{
  Examples of running Alg.~\ref{alg:dual-group} to identify the SDC machine, where $m=4, n=6$.
  H$i$ denotes group $i$ in the horizontal grouping phase, while V$i$ denotes group $i$ in the vertical grouping phase.
  }
  }
  \label{fig:replay}
  \vspace{-5mm}
\end{figure}

\noindent \textbf{Dual-Phase Replay.}
If training still fails, \oursys{} assumes unknown faults (\eg, SDCs) and resorts to group testing in a controlled setting for localization.
\camera{
For large-scale 3D parallel training~\cite{megatron2021}, machine stress testing and benchmarking~\cite{superbench} disrupt the original computing-communication pattern and data dependence of current LLM job~\cite{aegis}, undermining reproducibility.
}
To preserve diagnostic fidelity, we introduce a dual-phase, dimension-aware replay that keeps the original TP/PP sizes fixed while varying only the DP sizes (step $\textcircled{8}$).
Algorithm~\ref{alg:dual-group} details the localization procedure.
We partition machines into horizontal and vertical groups, \camera{reduce the model layers}, and replay the job on each group with a reduced DP size
(lines 2–8).
The intersection of the faulty horizontal and vertical groups pinpoints the failed machine (lines 9–11), which is then evicted (step $\textcircled{9}$).
In practice, we set $m = k \cdot$ \textsf{PP\_size}, $n=\textsf{DP\_size}/k$ where $k \in \mathbb{N}^+$, and $m \leq n$ for a unique solution.
Since \textsf{PP\_size} $\ll$ \textsf{DP\_size}, intra‑group communication remains representative.
\camera{
\camera{
As the depicted example in Fig.~\ref{fig:replay}, by replaying the training job twice and identifying the faulty group in each phase, SDC machine \#13 is isolated correctly.
This design choice effectively reduces unproductive time without the reliance on advanced diagnosis tools.
In our experience, each SDC incident typically involves only a single faulty machine, which is the common case in large-scale training.
}
}



\begin{figure*}[!t]
  \centering
  \includegraphics[width=\linewidth]{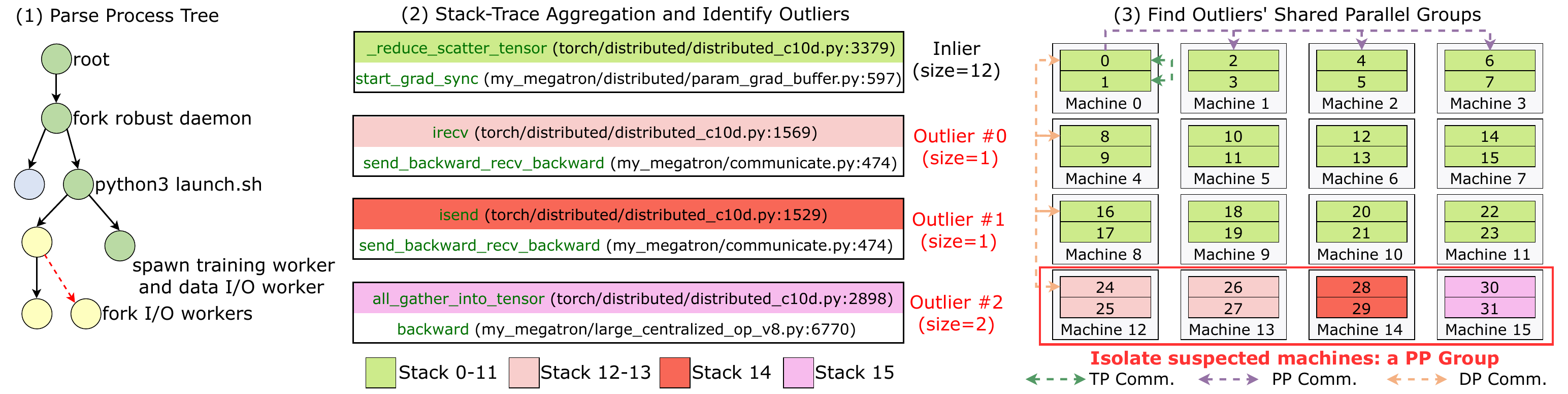}
  \caption{
  Stack aggregation for backward-communication hang pinpointing.
  The parallelism configuration: TP=2, PP=4, DP=4.
  }
  \label{fig:aggregate_stack}
\end{figure*}

\noindent \camera{\textbf{Lesson: Simple approaches address most incidents.}}
Based on empirical observations across 19 large-scale LLM training jobs ($\geq$ 9,600 GPUs), we find that direct machine eviction via real-time checks resolved 32.52\% of failures, reattempts recovered 22.70\%, and rollbacks handled another 9.20\%.
Only 1.23\% of failures required dual-phase replay.

\subsection{Case Study}\label{sec:autoFT_case_study}
\noindent \textbf{NaN loss diagnosis.}
In case that NaN losses are detected by the monitor during training,  
standard GPU and network tests are conducted first, including 
EUD and NCCL tests.
If all these tests are passed, a bit-wise alignment test is run:
\camera{
each machine initiates a reference model whose structure matches that of the target training job (\eg, dense models~\cite{gpt3} or MoE models~\cite{moe-layer}).
It loads predefined weights, employs a specific parallelism configuration (\eg, TP=2, PP=2, DP=2 or EP=2, PP=2, DP=2), and executes one training step on fixed input to ensure reproducibility.
}
The outputs from all machines are collected and analyzed to verify bit-wise accuracy.
Machines that yield incorrect results are promptly isolated and removed.
If this test does not identify any defective machines, reattempt and rollback
are sequentially employed to settle potential transient failures and human errors.
If training still fails, the dual-phase replay testing is applied for troubleshooting.



%

\section{Data-Driven Over-Eviction}
\label{sec:analyzer}
\camera{
Beyond NaN values, implicit failures also manifest as job hangs and MFU declines (Table \ref{tab:incident_stat}).
When a job gets hung, no traceable information is logged for stop-time diagnostics.
}
As for the decline in MFU, although the task slows down, all machines slow down simultaneously, and the throughputs of IO and RDMA, among others, all decline at the same time. All of these factors make it extremely difficult to identify the potentially faulty machines by analyzing the existing external information, as mentioned earlier. 
\camera{
To overcome this challenge, \oursys{} hooks into and inspects the stack traces of all internal training processes upon detecting these silent failures to localize faulty machines.
}
When receiving the aggregation triggering message, the controller notifies the on-demand tracer to capture process stack-traces, which are then sent to the runtime analyzer to conduct aggregation analysis in the background.
\camera{We first introduce the aggregation analysis mechanism through a job hang example, then present a case study for further illustration.}


\subsection{Aggregation Analysis}
\label{sec:agg_analysis}

To pinpoint the location of anomalies, the aggregation analysis compares the invocation stacks of GPU ranks in different machines.
\noindent Compared to events, training metrics, and \texttt{stdout/stderr} logs, stack-traces of processes offer a rich source of information for addressing complex incidents.
However, root causes may reside in subprocesses spawned by the main training processes for tasks like data fetching or checkpointing \cite{bytecheckpoint}; merely analyzing the stacks from the primary training processes is insufficient.
\camera{
\oursys{} incorporates this observation and performs a three-step aggregation to conduct a comprehensive analysis, based on the assumption that most healthy machines exhibit identical stack traces under a single implicit failure.
}

\camera{
Fig.~\ref{fig:aggregate_stack} illustrates a silent \camera{backward-communication} hang.
In this example, machine 15, which hosts the last stage of a model pipeline to generate activation gradients for backward propagation, stalls in \texttt{all\_gather\_into\_tensor}.
Meanwhile, unlike machines 0-11, which complete launching all backward-related kernels and proceed to gradient synchronization in the optimizer, machine 14 and machines 12-13 are blocked in \texttt{isend} and \texttt{irecv}, respectively, while transmitting gradients for certain micro-batches.
Conventional diagnostics make it difficult to efficiently and precisely determine the set of faulty machines.
Instead, \oursys{} addresses this issue by over-evicting isolated machines via a three-step procedure, avoiding the need to pinpoint the exact root cause.
}
First, \oursys{} parses the process trees in each training pod to identify training-related processes, \eg, \texttt{torchrun}, dataloader, and checkpoint processes.
\camera{
Next, stack-traces from these identified processes are aggregated into multiple groups via string matching to differentiate abnormal sources.
The dominant groups are deemed healthy (green stacks in Figure~\ref{fig:aggregate_stack}), while the remaining groups are classified as outliers (other colors).
}
Finally, we find the shared parallel groups for those outliers and isolate the corresponding machines.
In this example, the shared parallel group is one PP group (machines 12, 13, 14, and 15).
\camera{
The robust controller evicts the suspects and then resumes training.
}


\camera{
For fail-slow incidents (i.e., MFU decline), \oursys{} repeats aggregation every 10 seconds, flagging the parallel group with the most outliers at each round. The parallel group with the highest cumulative flag count across 5 rounds is marked as the degrader for over-eviction.
}


\subsection{Case Study}
\label{sec:case_studies}
Next, we dive into details of a representative implicit failure and how aggregation analysis works.


\noindent \textbf{Evaluation hang.}
We experienced task hangs during the LLM evaluation step~\cite{mmlu}, which typically measures model's multitask capacity.
In one example, the stack aggregation analysis isolated a specific pipeline spanning 6 machines, where the stacks in the intermediate stages differed from those of other ranks in the same DP$\times$TP group.
Those 
stages were hence stuck in their P2P communication operations.
The 6 
machines were automatically blacklisted and evicted, and the warm standby instances were scheduled 
for rapid replacement and training restarting.
Through background stress testing spanning several days, we finally determined the root cause: two of the machines have defective CUDA cores, causing hangs and preventing the P2P operations.

\section{\camera{Controlled and Swift Recovery}}
\label{sec:fast-recovery}
\sosp{
After failure detection and localization, \oursys{} restarts training swiftly in a consistent environment, minimizing downtime and avoiding new faults. 
Specifically, we apply in-place hot-update (Sec.~\ref{sec:hot_update}) for code/data adjustments, use warm standbys (Sec.~\ref{sec:warm_standby}) to eliminate scheduling costs, and employ \camera{over-eviction}‑aware checkpointing for fast snapshots and local safe backups (Sec.~\ref{sec:checkpoint}).  
}

\subsection{In-Place Hot-Update}
\label{sec:hot_update}

Manual training restarts for code adjustments are the norm during LLM training.
\sosp{
Rescheduling new machines for code upgrades or rollbacks not only incurs significant overheads but also introduces potentially faulty machines, complicating fault localization when failures occur after restarts.
To minimize the overhead and avoid the risks of deploying potentially faulty machines during restart,} a lazy hot-update mechanism is introduced for \textit{in-place} code modifications without destroying the existing pod environments.
Update strategies are tailored according to the nature of the code modification.
For urgent requests like bug fixes, training is immediately halted to apply the updates.
For less critical changes such as experimenting with new optimizations or updating software versions, updates are 
integrated into the recovery procedure upon the next failure, leveraging frequent interruptions observed in large-scale LLM training (\eg, interruptions occur on average once every 2.78 hours during Llama 3.1 training
~\cite{llama3.1}). In any case, a non-applied non-critical update is performed when a default triggering window (\eg, 24 hours) expires.
\sosp{All modifications will be persisted in our database, making them traceable and reproducible.}
The hot-update mechanism also makes the continuous integration of the evolving training code part of the pipeline in robust LLM training, through automatic apply and rollback (Sec.~\ref{sec:effective_resolution}).

\subsection{Warm Standby Machines}
\label{sec:warm_standby}

Whenever machine evictions occur, \oursys{} utilizes warm standby instances to quickly replace the missing machines for training resumption.
\camera{
Though introducing some GPU idling on standby machines, the reduced restart costs translate into the utilization improvements on healthy machines in the cluster,
especially under high-frequency training interruptions during large-scale training~\cite{llama3.1}.
}
We maintain a standby machine pool and decide the number of backup machines based on the key observation that failures in large-scale training are typically independent, happening at single nodes, and simultaneous failures involving multiple nodes are extremely rare
~\cite{opt, jit-check}.
We estimate the daily failure rate of individual machines 
using historical data 
and model simultaneous failures across machines 
by a binomial distribution.
We set the number of warm standby instances 
to the 99th percentile (P99) of this distribution, which can effectively meet the needs in most scenarios.

The standby machine pool is replenished dynamically.
Pod environment initialization is performed on each new \camera{standby} machine, including machine \camera{self-checks to ensure its healthy status}, image installation, and library downloading, which then enters a low-power sleep mode. 
Upon machine evictions, if there are sufficient standby
machines, they are directly awakened and integrated into training; otherwise, prompt replenishment is carried out, and training restarts when all needed machines finish their pod environment initialization.
\camera{An additional benefit of this design is that only a minimal number of machines change in the event of a failure and subsequent job restart; the remainder continue exactly as before, improving both resource efficiency and the controllability of model training.}


\subsection{Over-Eviction-Aware Checkpointing}
\label{sec:checkpoint}

\begin{figure}[!t]
  \centering
  \includegraphics[width=\linewidth]{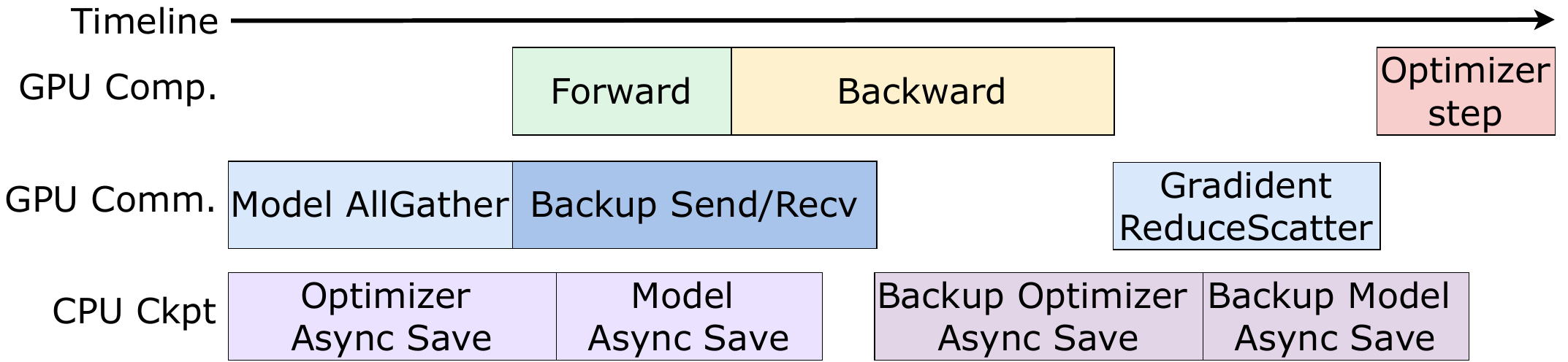}
  \caption{
  An example of checkpointing and backup operations scheduling with ZeRO-style parallelism.
  }
  \label{fig:ckpt}
\end{figure}

\oursys{} advocates in-memory checkpointing by saving and backing up checkpoints on both local and \camera{peer} machines.
\camera{
It employs a hierarchical checkpointing scheme that leverages host CPU memory and SSD storage tiers, incorporating a backup strategy that anticipates machine over-eviction (Sec.~\ref{sec:analyzer}) to guarantee availability.
By eliminating reliance on remote storage services over low-bandwidth frontend networks~\cite{check-n-run, bytecheckpoint}, \oursys{} prevents potential training hangs or crashes caused by storage-service failures (see Table~\ref{tab:incident_stat}, where we had 1104 HDFS errors).
}

\noindent \textbf{Operation scheduling.}
Through meticulous operation scheduling, \oursys{} achieves near-zero-overhead in-memory checkpointing. 
As the example in Fig.~\ref{fig:ckpt}, 
to back up sharded model and optimizer states, 
\oursys{} exploits the idle communication cycles in each training step, \ie, during forward and backward computations, and employs 
P2P communication for each rank to exchange these shards with its peer rank in selected backup machines (see details in the backup strategy).
These backup shards are then saved into CPU memory. 
Checkpoint I/O operations are performed in an asynchronous manner with forward and backward computation.
The optimizer step of GPU computation waits for the completion of saving each rank's own checkpoints to ensure data integrity.
Backup checkpoints can be saved concurrently with model and optimizer updates.
\oursys{} creates a separated CUDA stream to isolate the execution of training-related and checkpointing-related kernels.
For the backup communication executed in parallel with forward and backward propagations, we partition the states into small chunks, interleaving the transmission with training communication traffic in different parallelism dimensions.

\begin{figure}[!t]
  \centering
  \includegraphics[width=\linewidth]{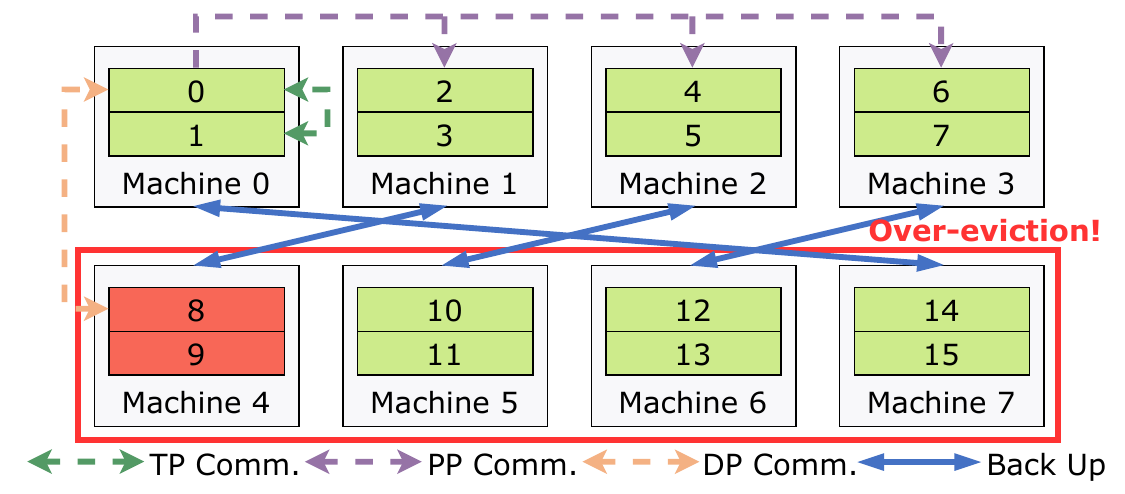}
  \caption{
  Checkpoint backup with over-eviction awareness.
  3D Parallelism configuration: TP=2, PP=4, DP=2.
  }
  \label{fig:ckpt_backup}
\end{figure}

\noindent \textbf{Cross parallel group backup strategy.}
Replicating sharded optimizer and model states across machines is essential to tolerate machine failures.
\camera{In \oursys{}, machine over-eviction mainly stems from aggregation analysis (Sec.~\ref{sec:analyzer}), where an entire parallel group can be evicted for prompt training restarts.}
Therefore, it is crucial to select target machines for safely storing backups upon over-eviction.
\oursys{} advocates a \textit{cross parallel group} backup strategy to tackle potential machine over-eviction.
As depicted in Fig.~\ref{fig:ckpt_backup}, during large-scale 3D parallel training, each rank backs up its sharded optimizer states outside of its 3D parallel groups.
For instance, ranks 8 and 9 exchange their optimizer states with ranks 2 and 3, ensuring that none share the same PP, DP, or TP groups.
Similarly, sharded model states, which are \camera{deduplicated within the DP group~\cite{bytecheckpoint}}, adhere to this backup strategy.
If the parallelism strategy comprises only a single parallel group (\eg, ZeRO parallelism), the system defaults to backup in neighboring machines.

\section{Implementation}
\label{sec:impl}


\noindent\textbf{Robust Controller and Agent.}
The Robust Controller consists of an orchestration module and a control module written in 20k lines in Golang.
We implement the orchestration module using Kubernetes Custom Resource Definitions (CRDs) to represent job operations 
(\textasciitilde 3k LoC). 
Each job has a runtime CRD for elastic training rules and a job CRD for pod scheduling. To enhance cluster management efficiency, we replace the standard etcd~\cite{etcd} with an internal metadata system and utilize an internal scheduler for pod group scheduling.
The control module centers around a job manager service that maintains job controllers (\textasciitilde 17k LoC). For each job, we register a dedicated controller service with the job manager, using goroutines to enable efficient resource sharing, various warm backup strategies, and unified failure recovery.
The Robust agent is a Python daemon(\textasciitilde 5k LoC) running alongside each job to manage training processes. The agent communicates with the controller through gRPC-based heartbeats and supports runtime hot updates.


\noindent\textbf{Runtime Analyzer}
is implemented in around 12k lines of Golang.
The analyzer standardizes anomalies by aggregating logs, I/O operations, host anomalies, on-demand tracer output, and pod anomalies into unified events. Due to the heterogeneous and dispersed nature of 
observed data and the need for rapid issue classification, we designed an event-driven system for real-time analysis.
It allows swift root cause localization and collaborates with the robust controller for fast failure handling.
For NCCL timeout issues, it collects stack traces from the on-demand tracer, which is implemented using py-spy~\cite{py-spy} and \camera{flight-recorder~\cite{flight-recorder}}, and training topology information to
facilitate troubleshooting. 
The analyzer also constructs a process tree of worker training processes, to meet various analysis needs.


\noindent\textbf{Warm Standby.}
We leverage the Robust Controller's orchestration module to maintain a specified number of warm standby nodes through asynchronous provisioning. Upon initialization, each Robust Agent queries the Controller to determine its state, either warm standby or active training. 
When execution reaches \sosp{pre-set} barriers that block code execution, processes on standby machines verify their current state. 
If they are in the standby state, these processes enter a polling loop, periodically checking with the Robust Controller for an activation signal. 
Once activation is signaled, training seamlessly resumes beyond the barrier, integrating warm standby nodes into the ongoing training workflow without disruption.

\noindent\textbf{High-Frequency Checkpointing}
is implemented in 3k lines of Python
with a dual-buffer 
for CPU tensors, which alternately stores the optimizer's state dictionary between different iterations.
We implement asynchronous checkpointing 
by overlapping three operations: Device-to-Host (D2H) copying, serialization, and sending to other ranks for backup.
When the first CPU tensor is undergoing D2H copying, we perform serialization or sending on the second tensor simultaneously. D2H operations are executed on a dedicated CUDA stream, enabling independent execution of D2H memory copy alongside training computation. Failure recovery is implemented by selecting the latest available checkpoints based on D2H and serialization completion status.

\section{Evaluation}
\label{sec:eval}
\sosp{
We first present deployment results to demonstrate how \oursys{} achieves robust training in real production (Sec.~\ref{sec:deployment}).
We then compare \oursys{} with several baselines to underscore its enhancements in failure recovery (Sec.~\ref{sec:failure_recovery}).
}

\noindent \textbf{Testbed.}
All experiments are conducted on production GPU clusters.
For deployment results in Sec.~\ref{sec:deployment}, up to 1200 machines are employed, each equipped with 8 NVIDIA Hopper 80GB GPUs. 
For evaluation in Sec.~\ref{sec:failure_recovery}, we use a total of 1024 machines, each equipped with 16 NVIDIA L20 48GB GPUs connected via 30GB/s PCIe, resulting in over 16,384 GPUs in total. 
All machines mentioned above are interconnected through eight 400 Gbps RDMA links and powered by 96-core Intel Xeon processors and equipped with 2 TB of DRAM.



\subsection{Robustness in Real Production}
\label{sec:deployment}

\camera{
\oursys{} has been deployed on \ourcompany{}'s production clusters to serve LLM training tasks.
}
We show that \oursys{} can effectively reduce incident detection time (Sec.~\ref{sec:exp_inspec}) and resolve incidents via the automatic fault tolerance framework and aggregation analysis (Sec.~\ref{sec:effective_resolution}).
The overall ETTR and MFU statistics are also reported to verify the end-to-end effectiveness (Sec.~\ref{sec:deploy_performance}).
Finally, we compare our automated fault tolerance framework with prior practice to justify its advantages (Sec.~\ref{sec:deploy_ablation}).
We collected two \camera{in-house pretraining jobs of our production-grade models}: a three-month job for training a dense model \camera{(Llama-like~\cite{llama3.1}, 70+B)}, and a one-month job for training an MoE model~\cite{doubao-1.5} \camera{(200+B)}.
These pretraining jobs were run on a GPU cluster comprising 9,600 Hopper GPUs.

\subsubsection{Reduce Detection Time}
\label{sec:exp_inspec}

\begin{table}[!t]
\caption{
The time to detect different 
infrastructure failures.
$T_{timeout}$ is the default timeout threshold of PyTorch-Distributed~\cite{pytorch-distributed} (\textasciitilde 10 minutes).
$T_{monitor}$ is the time interval to monitor the MFU decline.
}
\label{tab:infra_detect_time}
\resizebox{\linewidth}{!}{
\begin{tabular}{cccc}
\toprule
\textbf{Category} & \textbf{Root Cause} & \textbf{w/ Inspection (s)} & \textbf{w/o Inspection} \\
\midrule
\multirow{3}{*}{Network} & NIC crash & 30 & $T_{timeout}$ \\
& Port Flapping & 30 & $T_{timeout}$ \\
& Switch Down & \sosp{30 $\cdot$ 2} & $T_{timeout}$ \\
\midrule
\multirow{3}{*}{GPU} & Driver Hang & 10 & $T_{timeout}$ \\
& High Temperature & 10 & $T_{monitor}$ \\
& GPU Lost & 10 & $T_{timeout}$ \\
\midrule
\multirow{1}{*}{Host} & OS Kernel Fault & 2 & $T_{timeout}$ \\
\bottomrule
\end{tabular}}
\end{table}


We show that the real-time check mechanism effectively reduces failure detection time by comparing it with a baseline approach relying on only the timeout threshold   (\textasciitilde 30 minutes) and performance metric alerts from the monitor.
The alert frequency depends on actual training iteration time (Sec.~\ref{sec:proactive_checks}).

We implement different detection frequencies and judgment criteria for various infrastructure components, as shown in Table~\ref{tab:infra_detect_time}.
For example, the inspection interval for network components is set to 30 seconds.
We wait for two consecutive unresponsive switch events before raising an alert. 
File system errors, especially those due to panic-level OS kernel faults, are detected promptly. 
For high temperature issues, our system detects individual GPU overheating within 10 seconds and correlates this with MFU degradation to verify gray failures caused by 
thermal throttling GPUs.
The baseline system can only detect MFU decline after \sosp{gathering statistics} from multiple training iterations. 
By detecting failures earlier, we minimize unproductive idle periods and eliminate the need for stop-time diagnostics.


\subsubsection{Resolve Incident}
\label{sec:effective_resolution}


\begin{table}[!t]
\caption{
Distribution of resolved incidents across different mechanisms in two production jobs. 
Numbers represent incident counts, with percentages shown in parentheses.
}
\label{tab:aggregation_benfits}
\resizebox{\linewidth}{!}{
\begin{tabular}{c|c|c|c|c}
\toprule
\textbf{Job}                  & \textbf{Mechanism}          & \textbf{Explicit}  & \textbf{Implicit}  & \textbf{Manual Restart}      \\
\midrule
\multirow{3}{*}{Dense} & AutoFT-ER             & 128 (73.1\%)   & $\dagger$     & $\dagger$      \\
& AutoFT-HU & $\dagger$              & $\dagger$     & 20 (11.4\%) \\
& Analyzer-ER         & $\dagger$              & 15 (8.6\%)      & $\dagger$ \\
& Rollback & $\dagger$              & 12 (6.9\%)     & $\dagger$ \\
\midrule
\multirow{3}{*}{MoE}   & AutoFT-ER              & 71 (56.8\%)    & $\dagger$      & $\dagger$      \\
& AutoFT-HU & $\dagger$              & $\dagger$     & 31 (24.8\%) \\
& Analyzer-ER         & $\dagger$              & 9 (7.2\%)      & $\dagger$ \\
& Rollback & 1 (0.8\%)            & 13 (10.4\%)    & $\dagger$ \\
\bottomrule
\end{tabular}}
\end{table}

Table~\ref{tab:aggregation_benfits} presents 
incident resolution ratios from the two production jobs,
by four mechanisms in \oursys: 1) AutoFT-ER is the automated fault tolerance mechanism with machine eviction followed by training restart; 
2) AutoFT-HU represents the hot-update mechanism; 
3) Analyzer-ER corresponds to incidents diagnosed by the aggregation analyzer and resolved through machine eviction and training restart; and 4) Rollback reverts the code to a previous stable version. 
The majority of explicit failures were resolved by automatic machine eviction and training restart, accounting for 73.1\% and 56.8\% of incidents in the two jobs, respectively.
\camera{
For implicit failures (job hangs and MFU declines in the two jobs),
the analyzer successfully resolved 24 incidents through machine over-eviction, avoiding the need for human intervention, significantly reducing the unproductive time.
In addition, rollback identified several engineering code issues, accounting for 6.9\% and 11.2\% of incidents in the two jobs, respectively.
}
Finally, we observe that all manual restart requirements related to code and data adjustments are handled by the hot-update mechanism.



\subsubsection{Guarantee Performance}
\label{sec:deploy_performance}

\begin{figure}[!t]
  \centering
  \includegraphics[width=\linewidth]{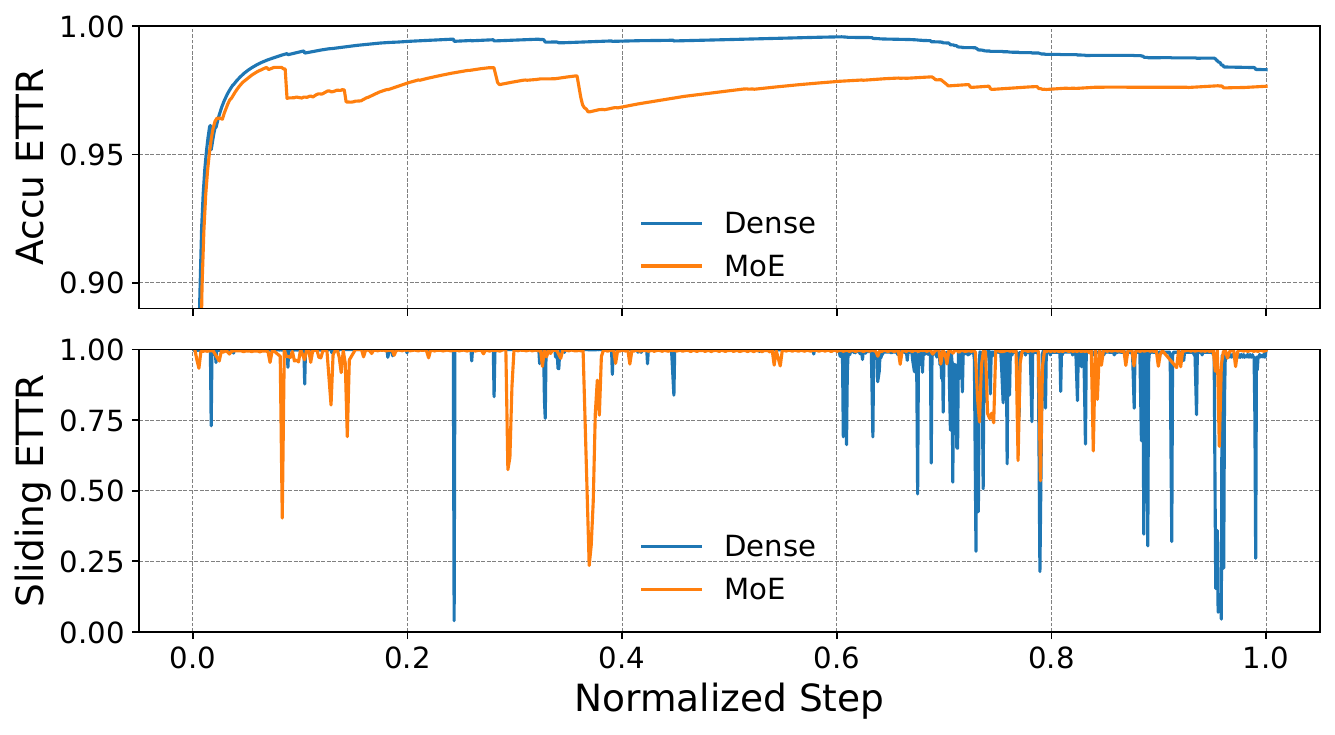}
  \caption{Cumulative ETTR and sliding-window ETTR in dense LLM and MoE pre-training jobs. 
  }
  \label{fig:deploy_ettr}
\end{figure}

\begin{figure}[!t]
  \centering
  \includegraphics[width=\linewidth]{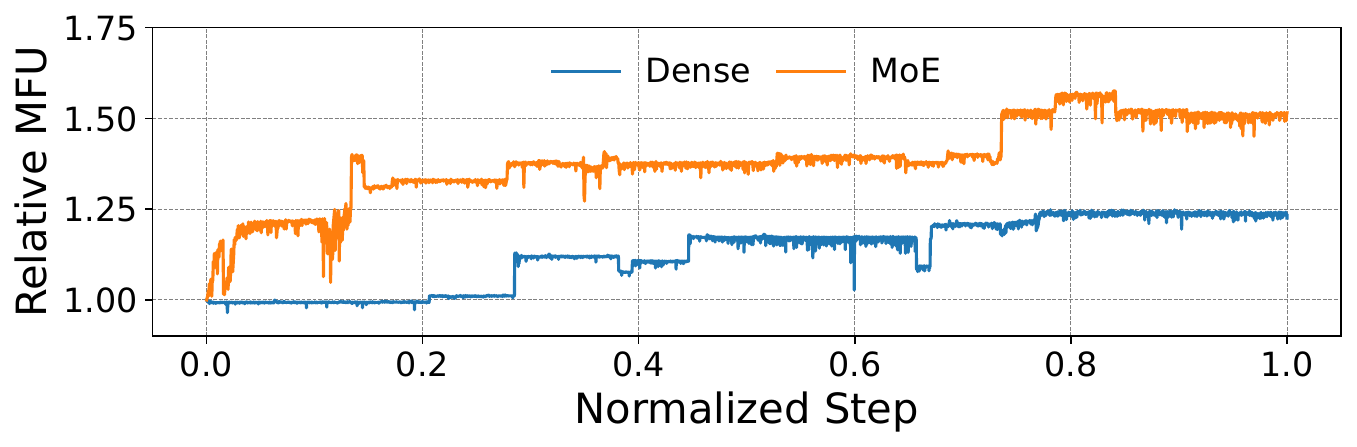}
  \caption{Relative MFU from dense LLM/MoE training jobs, as the ratio 
  to the respective minimum MFU value.}
  \label{fig:deploy_mfu}
\end{figure}

\sosp{
For performance evaluation, we measure ETTR and MFU on both Dense and MoE LLM training jobs.
}
We define \textit{Cumulative ETTR}, which is computed as the ratio of the accumulated productive training time to the cumulated wall-clock time of a job run~\cite{meta-cluster}.
\sosp{
However, for long‑running jobs, this aggregate metric obscures the temporal dynamics of failure handling and recovery.
We introduce \textit{sliding‑window ETTR}, computed over a one‑hour window, which more accurately reflects the impact of intermittent failures.
Results are depicted in Fig.~\ref{fig:deploy_ettr} and Fig.~\ref{fig:deploy_mfu}.
}

\camera{
\oursys{} maintained the cumulative ETTR at a plateau of up to 97\% and kept the unproductive training time within a maximum of 50 minutes for both jobs.
}
We observed in Fig.~\ref{fig:deploy_ettr} that during the later training periods of the two models, the cumulative ETTR experienced slight degradation while the sliding-window ETTR exhibited increased fluctuations.
These changes are caused by the following: The engineering team deployed a feature for long-context training for the first time, which led to several code failures; then, the prolonged training time made the cluster more vulnerable, causing more frequent performance degradation and failures.
Despite the increased frequency of failures and manual restarts, \oursys{} was able to detect, diagnose, and recover the training status efficiently.

\camera{
The relative MFU of the two jobs kept increasing as training progressed.
}
During training, we initially deployed a naive version of the pretraining code on the cluster, and then continuously tuned and optimized its learning process and computational efficiency.
In Fig.~\ref{fig:deploy_mfu}, each leap in the MFU curves indicates that a more efficient version of the training code was deployed through \oursys{}'s hot update, which caused only a negligible degradation in ETTR.
We achieved a 1.25$\times$ and 1.58$\times$ MFU improvement as compared to the initial run in the dense and MoE jobs, respectively.

\camera{
We also observed that the ETTR of MoE training is relatively lower compared to dense models (Fig.~\ref{fig:deploy_ettr}).
Unlike the training of dense models, whose performance is typically well-optimized by the community~\cite{megatron2019, megatron2021, megatron2023, megascale}, MoE training often involves the integration of numerous custom optimizations, such as GPU kernel tuning, computation-communication overlapping, and load balancing strategies.
While these optimizations are necessary for improving training efficiency, showing higher MFU (Fig.~\ref{fig:deploy_mfu}), they also introduce additional complexity, increasing the likelihood of rollbacks and manual restarts.
}

\begin{table}[!t]
\caption{
Training setup of two sparse LLM training jobs. Scale gives the number of machines $\times$ GPUs used for training. P99 indicates the number of backup machines $\times$ GPUs.
Catastrophic represents the number of machines $\times$ GPUs involved in extreme failure cases 
(< 1\% probability).
}
\label{tab:model_parallel_config}
\resizebox{\linewidth}{!}{
\begin{tabular}{cccccc}
\toprule
\textbf{Model} & \textbf{Scale} & \textbf{Parallelism} & \textbf{Batch Size} & \textbf{\#P99} & \textbf{\#{Catastrophic}}\\
\midrule
\multirow{2}{*}{70B} & 128$\times$16 & TP=8, DP=32, PP=8 & 512 & 2$\times$16 & 32$\times$16 \\
& 256$\times$16 & TP=8, DP=64, PP=8 & 1024 & 2$\times$16 & 32$\times$16 \\
\midrule
\multirow{2}{*}{256B} & 512$\times$16 & TP=8, DP=64, PP=16 & 1024 & 3$\times$16 & 32$\times$16 \\
& 1024$\times$16 & TP=8, DP=128, PP=16 & 2048 & 4$\times$16 & 32$\times$16 \\ 
\bottomrule
\end{tabular}}
\end{table}

\begin{table}[]
\caption{
Incident resolution cost comparison.
}
\label{tab:error}
\resizebox{\linewidth}{!}{
\begin{tabular}{c|c|c|c}
\toprule
\textbf{Incident Symptoms}                      & \textbf{Ours Mean (s)} & \textbf{Ours Max (s)} & \textbf{Selective (s)} \\
\midrule
CUDA Error                   & 93     & 600    & 518 (INF)     \\
Inifiband Error & 60     & 60     & 288           \\
HDFS Error                   & 58     & 65     & INF              \\
OS Kernel Panic              & 109    & 120    & 168           \\
GPU Memory Error             & 10     & 10     & 600           \\
NaN Value                    & 4289   & 7200   & 7200 (INF)       \\
GPU Unavailable              & 10     & 10     & 120           \\
Code/Data Adjustment         & 57     & 64     & INF              \\
\bottomrule
\end{tabular}}
\end{table}

\subsubsection{Compare with Prior Practice}
\label{sec:deploy_ablation}
\sosp{
We collect incident symptoms, logs, and exit codes from Dense/MoE jobs and compare our automated fault‑tolerance framework against selective stress testing~\cite{superbench, characterization}, one of the most common practices for troubleshooting in previous works~\cite{superbench, characterization}.
For \oursys{}, we measure the time from failure localization to successful restart.
\camera{
For the baseline method, we conduct corresponding
stress testing (\eg, GPU-related, network-related, \etc) guided by indicators in logs and exit codes, recording the testing time.
}
As Table~\ref{tab:error} shows, our approach cuts mean resolution time across all symptoms, up to an 84.50\% reduction for CUDA errors.
For symptoms due to human mistakes, the baseline’s stress tests fail to localize the fault, whereas our rollback mechanism pinpoints and recovers from them.
}


\subsection{Efficiency in Failure Recovery}
\label{sec:failure_recovery}

\sosp{
We highlight the performance improvements in job restart (Sec.~\ref{sec:recovery_restart}) and checkpointing (Sec.~\ref{sec:recovery_checkpoint}).
achieved by \oursys{} through different techniques.
}

\subsubsection{Swift Job Restart}
\label{sec:recovery_restart}

\begin{table}[!t]
\caption{
Scheduling time comparison between requeue and hot update mechanisms upon 5 code update events.
}
\label{tab:schedule_time_update}
\resizebox{\linewidth}{!}{
\begin{tabular}{c|c|c|c|c}
\toprule
\textbf{Scale (\# GPUs)}      & \textbf{128$\times$16} & \textbf{256$\times$16} & \textbf{512$\times$16} & \textbf{1024$\times$16} \\
\midrule
Requeue (s)       & 454
 & 545 & 635 & 768  \\
Hot updates (s) & 46  & 51  & 54  & 65  \\
\bottomrule
\end{tabular}}
\end{table}

\begin{figure}[!t]
  \centering
  \includegraphics[width=0.95\linewidth]{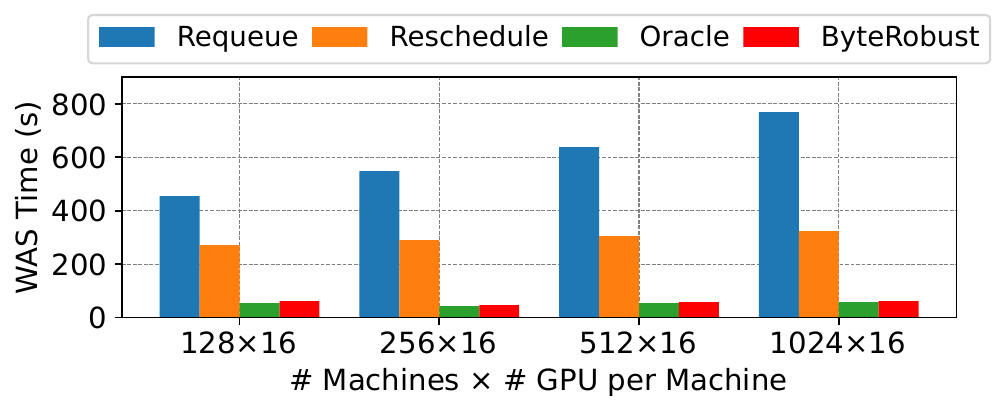}
  \vspace{-3mm}
  \caption{
Weighted average scheduling (WAS) time 
upon machine eviction events. 
  }
  \label{fig:weighted_schedule_time_evict}
\end{figure}

We compare \camera{\oursys{}'s job restart mechanism} with three baselines:
(i) \textit{requeue}~\cite{meta-cluster, kubedl, kubeflow, vcjob}: kill and requeue the entire job, reallocating all machines;
(ii) \textit{reschedule}\cite{pathways}: spin up replacements only for evicted machines, reinstalling pods on them;
(iii) \textit{oracle}: assume unlimited warm standbys ready to replace any evicted machine.

\noindent \textit{Ours}: we provision warm standbys based on the P99 failure count (4 backups for a 1,024‑instance job). If evictions $\leq 4$, we reuse standbys and replenish them asynchronously; if evictions $\geq 4$, we only reschedule the shortfall.


\noindent \textit{Efficient hot-update.}
We measure scheduling time (from the update request to job resume) across five manual code‐change events.
By reusing the existing environment, our hot‐update mechanism is $11.04\times$ faster than full requeue (Table~\ref{tab:schedule_time_update}).


\noindent \textit{Efficient warm standby.}
For each training scale, we first identify the P99 faulty‑machine count N, then simulate evictions of 1 to $N$ machines, plus catastrophic switch failures (all machines evicted, \eg, 32 nodes).
We measure scheduling time from failure detection to job resume and compute a weighted‑average scheduling time (WAS) by weighting each scenario according to a binomial distribution (Sec.\ref{sec:warm_standby}), with catastrophic failures fixed at 1\%.
As Fig.\ref{fig:weighted_schedule_time_evict} shows, our warm‑standby approach reduces recovery time by $10.87\times$ versus requeue and $5.36\times$ versus reschedule, and is within just $5.19\%$ of the oracle upper bound.


\noindent \textit{Scalability.}
From Fig.~\ref{fig:weighted_schedule_time_evict} we find that as the training scale grows, requeue’s restart time increases markedly.
This is due to the costs of clear job metadata, reallocate instance quotas, reinstall, and reset the entire pod environment.
On the contrary, our warm standby and hot‐update incur constant, low overhead—demonstrating superior scalability.

\subsubsection{Near Zero-Overhead Checkpointing}
\label{sec:recovery_checkpoint}

\begin{table}[!t]
\caption{
Checkpointing efficiency comparison. 
MFU values are relative to the training MFU without checkpointing.
}
\label{tab:ckpt_efficiency}
\resizebox{\linewidth}{!}{
\begin{tabular}{c|cc|cc}
\toprule
\textbf{Model} & \textbf{Scale} & \textbf{Approach} & \textbf{Blocking Time (s)} & \textbf{MFU (\%)}\\
\midrule
\multirow{6}{*}{70B} & \multirow{3}{*}{128$\times$16} & Megatron save & 6.77 & 39.84 \\
& & Memory save & 1.84 & 70.05\\
& & \oursys{} save & \textbf{0.04} & \textbf{99.23}\\
\cline{2-5}
& \multirow{3}{*}{256$\times$16} & Megatron save & 7.14 & 39.11 \\
& & Memory save & 1.69 & 72.36\\
& & \oursys{} save & \textbf{0.03} & \textbf{99.12}\\
\midrule
\multirow{6}{*}{256B} & \multirow{3}{*}{512$\times$16} & Megatron save & 13.02 & 43.07 \\
& & Memory save & 0.22 & 95.90 \\
& & \oursys{} save & \textbf{0.01} & \textbf{99.71} \\
\cline{2-5}
& \multirow{3}{*}{1024$\times$16} & Megatron save & 12.98&42.80 \\
& & Memory save &  0.18 & 96.92 \\
& & \oursys{} save &  \textbf{0.02} & \textbf{99.11}\\
\bottomrule
\end{tabular}}
\end{table}

We evaluate our checkpointing on sparse MoE~\cite{moe-layer} LLMs (70B and 256B model sizes), leveraging 3D parallelism~\cite{megatron2021} with ZeRO-1~\cite{zero} on real-world text generation tasks.
Detailed configurations are given in Table~\ref{tab:model_parallel_config}. 
We compare:
(i) \textit{\oursys{} save}, our module;
(ii) \textit{Memory save}, in-memory checkpointing, proposed by Gemini~\cite{aws-gemini};
(iii) \textit{Megatron save}, blocking checkpointing in Megatron‑LM~\cite{megatron2019}.
We evaluate the training blocking time (i.e., checkpoint stalls) and the MFU difference with and without enabling checkpointing during training.
The checkpointing frequency is set to one iteration.
As Table~\ref{tab:ckpt_efficiency} shows, \oursys{} save cuts blocking time by 99.69\% and 95.10\% versus \textit{Megatron save} and \textit{Memory save}.
By exploiting idle PCIe and network bandwidth in ZeRO‑style parallelism to overlap I/O with training, it limits MFU loss to just 0.71\%, improving over \textit{Megatron save} and \textit{Memory save} by 98.8\% and 89.6\%, respectively.
\section{Experiences and Limitations}
\label{sec:experience}

\noindent\textbf{Immature Diagnostic Tools.}
\sosp{
As GPU hardware evolves rapidly, associated monitoring and diagnostic tools often lag in maturity, making fault root cause analysis challenging. To ensure robustness under these constraints, we introduce system-level adaptations, including application-level isolation strategies such as data-driven over-eviction and dual-phase replay. These techniques are particularly valuable during the early deployment of GPU clusters with new hardware, though their necessity diminishes as diagnostic tools mature. Notably, diagnostic tools themselves can occasionally introduce new faults. In one case, we observed MFU degradation during training, traced to a diagnostic procedure (EUD) that inadvertently lifted a previously applied frequency lock, resulting in unexpected GPU downclocking. 
}

\vspace{-3mm}
\noindent\textbf{False Positive.}
\sosp{
False positives primarily stem from two sources. (i) Diagnostic tool limitations: Imperfections in tools such as EUD or network diagnostics can trigger erroneous alerts, leading to the unnecessary eviction and stress testing of healthy machines, with minor impact on cluster utilization. (ii) Intentional over-eviction: To expedite fault localization in 3D parallel training, we evict entire pipeline parallel (PP) groups (e.g., 8 machines per group in a 9600-GPU job), even though only 1–2 nodes are typically faulty. While this results in 6–7 false positives, the trade-off is acceptable given that training jobs commonly involve around 10,000 GPUs, and early isolation significantly reduces recovery time.
}

\noindent\textbf{Silent Data Corruption.}
\sosp{
SDC is a critical yet often overlooked challenge in scaling large language model (LLM) training to the next-scale. SDC results from factors such as input-sensitive numerical instabilities, race conditions, and thermal variations, causing incorrect computations like NaN values or gradient anomalies~\cite{sdcgoogle, llama3.1}. The collective communication patterns in distributed training exacerbate the propagation of these errors across multiple machines. Deep Learning's inherent robustness can obscure such faults, making them difficult to detect. In our production environment, NVIDIA's EUD diagnostic tool~\cite{eud} achieves only ~70\% recall. To mitigate this, we developed the MiniGPT verification suite, using deterministic workloads for intra-machine validation and dual-phase replay testing for inter-machine fault reproduction. However, these methods incur significant overhead, and as training scales increase, so does the frequency and impact of SDC, underscoring the need for more efficient detection, isolation, and diagnosis techniques.
}

\vspace{-4mm}
\section{Related Work}
\label{sec:related}

\noindent \textbf{Fault-Tolerant LLM Training.}
Megascale~\cite{megascale} combines 
periodical heartbeats and RDMA metrics monitoring to detect faults and run lightweight stop-time checks for diagnosis.
However, it cannot automatically isolate suspected machines when detecting anomalies in RDMA traffic, necessitating manual investigations.
Hu \textit{et al.}~\cite{characterization} incorporate an LLM-based log agent with rule-based heuristics to improve the accuracy of stop-time diagnostics. 
It relies on the log data and 
does not leverage runtime information, while the runtime data could enable faster and more precise identification of root causes, particularly in cases of implicit failures.
Both systems 
employ asynchronous checkpointing, but do not offer 
backup strategies~\cite{swift, aws-gemini}, while \oursys{} proposes a novel over-eviction-aware strategy.

\noindent \textbf{Elastic and Resilient Training.}
Other efforts have been put into enhancing training elasticity and resilience to prevent training interruptions~\cite{varuna21, easyscale, bamboo, oobleck, parcae, recycle, wagenlander2024tenplex}.
Bamboo~\cite{bamboo} introduces cross-stage redundant computations into the model pipeline, enabling continued training on spot instances.
Oobleck~\cite{oobleck} introduces the pipeline instantiation mechanism via predefined pipeline templates to tolerate concurrent failures in different pipelines.
Parcae~\cite{parcae} proactively adjusts parallelism strategies before instance preemptions and optimizes instance migration to achieve high liveput.
However, these methods are limited by certain parallelism strategies (\eg, DP, TP), while \oursys{} supports a wide range of prevalent parallelism strategies for LLM training.

\noindent \textbf{Gray Failure \sosp{and SDC}.}
A number of research studies have investigated characteristics of deep learning training jobs~\cite{weng2022mlaas, jeon2019analysis} and the failure incidents~\cite{gao2024, zhang2020an,characterization}.
Gray failure~\cite{grayfailure2017} is studied in-depth for storage systems~\cite{fail-slow, lu2023perseus, zhang2024msfrd}, data center networks~\cite{NetBouncer} and cloud services~\cite{incident2023}. 
However, the causes of gray failures in LLM training are rarely explored.
Ekko~\cite{sima2022ekko} routes requests to avoid fail-slow in parameter servers of deep learning recommendation systems.
{SuperBench}~\cite{superbench} introduces deep learning benchmarking to locate faulty GPU machines.
For large-scale LLM training, \oursys{} exploits runtime stack clustering and coarse-grained isolation to mitigate potential gray failures swiftly.
\sosp{
SDC is another major category of hard-to-detect faults, with recent research~\cite{sdcgoogle, sdcscale, sdcali} primarily focusing on its impact in CPU workloads. However, we observe that SDC also significantly affects large-scale LLM training on GPU clusters. While such faults are infrequent, their consequences—such as sudden loss spikes or NaN values—can be severe. These symptoms may overlap with data errors or engineering bugs, and SDC may not be deterministically reproducible. This complexity often results in prolonged fault diagnosis and resolution, ultimately limiting the scalability of LLM training.
}

\noindent \textbf{Checkpointing.}
Check-N-Run~\cite{check-n-run} employs differential checkpointing, which saves only the altered parts of the model and uses quantization to minimize the checkpoint size.
CheckFreq~\cite{checkfreq} pipelines training state saving with ongoing training to reduce stalls.
Gemini~\cite{aws-gemini} stores checkpoints in CPU memory with inter-machine backups, facilitating high-frequency checkpointing.
ByteCheckpoint~\cite{bytecheckpoint} unifies checkpoints from different training frameworks into a parallelism-agnostic representation, enabling efficient load-time resharding and high scalability. 
\oursys{} steps further to incorporate checkpointing with fine-grained scheduling and backup with eviction strategy awareness. 
\vspace{-2mm}
\section{Conclusion}
\label{sec:conclude}


We present \oursys{}, an LLM training management system deployed in \ourcompany{}'s GPU clusters. Drawing from extensive experience in large-scale LLM training, \oursys{} integrates fault characteristics, diagnostic capabilities, and LLM-specific features into a comprehensive system design. It employs an automated fault tolerance framework that efficiently distinguishes fault types, using runtime state analysis and data-driven methods to detect and isolate faulty machines.
We introduce mechanisms for efficient failover, including aggregated hot updates, warm backup machines, and fault-aware checkpointing, minimizing downtime.
Our insights aim to inspire further research and enhance the reliability of LLM training systems.
\vspace{-3mm}
\section{Acknowledgment}\label{sec:ack}
We thank our shepherd, Mahesh Balakrishnan, and the anonymous SOSP reviewers for their insightful feedback. We are grateful to Xiang Li and He Sun for implementing the initial version of the system. We also thank the ByteDance infrastructure and SRE teams.
This work was supported in part by a ByteDance collaborative research grant and grants from Hong Kong RGC under contracts 17204423, 17205824, and C7004-22G (CRF).

\bibliographystyle{ACM-Reference-Format}
\bibliography{ref}

\end{document}